# Controlling Complexity in Part-of-Speech Induction

**João V. Graça**                                                    JOAO.GRACA@L2F.INESC-ID.PT
*L²F INESC-ID*
*Lisboa, Portugal*

**Kuzman Ganchev**                                                      KUZMAN@GOOGLE.COM
*Google Inc.*
*New York, NY, USA*

**Luísa Coheur**                                               LUISA.COHEUR@L2F.INESC-ID.PT
*L²F INESC-ID*
*Lisboa, Portugal*

**Fernando Pereira**                                                    PEREIRA@GOOGLE.COM
*Google Inc.*
*Mountain View, CA, USA*

**Ben Taskar**                                                       TASKAR@CIS.UPENN.EDU
*Computer & Information Science*
*University of Pennsylvania*

## Abstract

We consider the problem of fully unsupervised learning of grammatical (part-of-speech) categories from unlabeled text. The standard maximum-likelihood hidden Markov model for this task performs poorly, because of its weak inductive bias and large model capacity. We address this problem by refining the model and modifying the learning objective to control its capacity via parametric and non-parametric constraints. Our approach enforces word-category association sparsity, adds morphological and orthographic features, and eliminates hard-to-estimate parameters for rare words. We develop an efficient learning algorithm that is not much more computationally intensive than standard training. We also provide an open-source implementation of the algorithm. Our experiments on five diverse languages (Bulgarian, Danish, English, Portuguese, Spanish) achieve significant improvements compared with previous methods for the same task.

## 1. Introduction

Part-of-speech (POS) categories are elementary building blocks for the syntactic analysis of text that play an important role in many natural-language-processing tasks, from machine translation to information extraction. While English and a handful of other languages are fortunate enough to have comprehensive POS-annotated corpora such as the Penn Treebank (Marcus, Marcinkiewicz, & Santorini, 1993), most of the worlds' languages have extremely limited linguistic resources. It is unrealistic to expect annotation efforts to catch up with the explosion of unlabeled electronic text anytime soon. This lack of supervised data will likely persist in the near future because of the investment required for accurate linguistic annotation: it took two years to annotate 4,000 sentences with syntactic parse trees for the Chinese Treebank (Hwa, Resnik, Weinberg, Cabezas, & Kolak, 2005) and four to seven years to annotate 50,000 sentences across a range of languages (Abeillé, 2003).

Supervised learning of taggers from POS-annotated training text is a well-studied task, with several methods achieving near-human tagging accuracy (Ratnaparkhi, 1996; Toutanova, Klein, Manning, & Singer, 2003; Shen, Satta, & Joshi, 2007). However, POS induction – where one does not have access to a labeled corpus – is a more difficult task with much room for improvement. In recent literature, POS induction has been used to refer to two different tasks. For the first one, in addition to raw text, we are given a dictionary containing the possible tags for each word and the goal is to disambiguate the tags of a particular word occurrence (Merialdo, 1994). For the second task, we are given raw text, but no dictionary is provided; the goal is to cluster words that have the same grammatical behavior. In this work, we target this latter, more challenging, unsupervised POS induction task.

Recent work on this task typically relies on distributional or morphological features, since words with the same grammatical function tend to occur in similar contexts and to have common morphology (Brown, deSouza, Mercer, Pietra, & Lai, 1992; Schütze, 1995; Clark, 2003). However, those statistical regularities are not enough to overcome several challenges. First, the algorithm has to decide how many clusters to use for broad syntactic categories (for instance, whether to distinguish between plural and singular nouns). Second, category size distribution tends to be uneven. For example, the vast majority of the word types are open class (nouns, verbs, adjectives), and even among open class categories, there are many more nouns than adjectives. This runs contrary to the learning biases in commonly-used statistical models. A common failure of those models is to clump several rare categories together and split common categories.

For individual word types, a third challenge arises from ambiguity in grammatical role and word sense. Many words can take on different POS tags in different occurrences, depending on the context of occurrence (the word *run* can be either a verb or a noun). Some approaches assume (for computational and statistical simplicity) that each word can only have one tag, aggregating all local contexts through distributional clustering (Schütze, 1995). While this one-tag-per-word assumption is clearly wrong, across many languages for which we have annotated corpora, such methods perform competitively with methods that can assign different tags to the same word in different contexts (Lamar, Maron, Johnson, & Bienenstock, 2010). This is partly due to the typical statistical dominance of one of the tags for a word, especially if the corpus includes a single genre, such as news. The other reason is that less restrictive models do not encode the useful bias that most words typically take on a very small number of tags.

Most approaches that do not make the one-tag-per-word assumption take the form of a hidden Markov model (HMM) where the hidden states represent word classes and the observations are word sequences (Brown et al., 1992; Johnson, 2007). Unfortunately, standard HMMs trained to maximize likelihood perform poorly, since the learned hidden classes do not align well with true POS tags. Besides the potential model estimation errors due to non-convex optimization involved in training, there is a more pernicious problem. Typical maxima of likelihood do not align well with maxima of POS tag accuracy (Smith & Eisner, 2005; Graça, Ganchev, Pereira, & Taskar, 2009), suggesting serious mismatch between model and data.

In this work, we significantly reduce that modeling mismatch by combining three ideas:

- The standard HMM treats words as atomic units, without using orthographic and morphological information. That information is critical to generalization in many languages (Clark, 2003). To address this problem, we reparameterize the standard HMM by replacing the multinomial emission distributions by maximum-entropy models (similar to the work of Berg-Kirkpatrick, Bouchard-Côté, DeNero, & Klein, 2010 and Graça, 2010). This allows the use of orthographic and morphological features in the emission model. Moreover, the standard HMM model has a very large number of parameters: the number of tags times the number of word types. This presents an extremely rich model space capable of fitting irrelevant correlations in the data. To address this problem we dramatically reduce the number of parameters of the model by discarding features with small support in the corpus, that is, those involving rare words or word parts.

- The HMM model allows a high level of ambiguity for the tags of each word. As a result, when maximizing the marginal likelihood, common words typically tend to be associated with every tag with some non-trivial probability (Johnson, 2007). However, a natural property of POS categories across many languages and annotation standards is that each word only has a small number of allowed tags. To address this problem we use the posterior regularization (PR) framework (Graça, Ganchev, & Taskar, 2007; Ganchev, Graça, Gillenwater, & Taskar, 2010) to constrain the ambiguity of word-tag associations via a sparsity-inducing penalty on the model posteriors (Graça et al., 2009).

We show that each of the proposed extensions improves the standard HMM performance, and moreover, that the gains are nearly additive. The improvements are significant across different metrics previously proposed for this task. For instance, for the 1-Many metric, our method attains a 10.4% average improvement over the regular HMM. We also compare the proposed method with eleven previously proposed approaches. For all languages but English and all metrics except 1-1, our method achieves the best published results. Furthermore, our method appears the most stable across different testing scenarios and always shows competitive results. Finally, we show how the induced tags can be used to improve the performance of a supervised POS tagging system in a limited labeled data scenario. Our open-source software for POS induction and evaluation is available at `http://code.google.com/p/pr-toolkit/`.

This paper is organized as follows. Section 2 describes the basic HMM for POS induction and its maximum-entropy extension. Section 3 describes standard EM and our sparsity-inducing estimation method. Section 4 presents a comprehensive survey of previous fully unsupervised POS induction methods. In Section 5 we provide a detailed experimental evaluation of our method. Finally, in Section 6, we summarize our results and suggest ideas for future work.

## 2. Models

The model for all our experiments is based on a first order HMM. We denote the sequence of words in a sentence as boldface $\mathbf{x}$ and the sequence of hidden states which correspond to part-of-speech tags as boldface $\mathbf{y}$. For a sentence of length $l$, we have thus $l$ hidden state variables $y_i \in \{1, \ldots, J\}, 1 \leq i \leq l$ where $J$ is the number of possible POS tags, and $l$ observation variables $x_i \in \{1, \ldots, V\}, 1 \leq i \leq l$, where $V$ is the number of word types. To simplify notation, we assume that every tag sequence is prefixed with the conventional start tag $y_0 = \texttt{start}$, allowing us to write as $p(y_1|y_0)$ the initial state probability of the HMM.

The probability of a sentence $\mathbf{x}$ along with a particular hidden state sequence $\mathbf{y}$ is given by:

$$p(\mathbf{x}, \mathbf{y}) = \prod_{i=1}^{l} p_t(y_i \mid y_{i-1}) p_o(x_i \mid y_i), \tag{1}$$

where $p_o(x_i \mid y_i)$ is the probability of observing word $x_i$ given that we are in state $y_i$ (emission probability), and $p_t(y_i \mid y_{i-1})$ is the probability of being in state $y_i$, given that the previous hidden state was $y_{i-1}$ (transition probability).

## 2.1 Multinomial Emission Model

Standard HMMs use multinomial emission and transition probabilities. That is, for a generic word $x_i$ and tag $y_i$, the observation probability $p_o(x_i \mid y_i)$ and the transition probability $p_t(y_i \mid y_{i-1})$ are multinomial distributions. In the experiments we refer to this model simply as the HMM. This model has a very large number of parameters because of the large number of word types (see Table 1). A common convention we follow is to lowercase words as well as to map words occurring only once in the corpus to a special token 'unk'.

## 2.2 Maximum Entropy Emission Model

In this work, we use a simple modification of the HMM model discussed in the previous section: we represent conditional probability distributions as maximum entropy (log-linear) models. Specifically, the emission probability is expressed as:

$$p_o(x|y) = \frac{\exp(\theta \cdot \mathbf{f}(x, y))}{\sum_{x'} \exp(\theta \cdot \mathbf{f}(x', y))} \tag{2}$$

where $\mathbf{f}(x, y)$ is a feature function, x ranges over all word types, and $\theta$ are the model parameters. We will refer to this model as HMM+ME. In addition to word identity, features include orthography- and morphology-inspired cues such as presence of capitalization, digits, and common suffixes. The feature sets are described in Section 5. The idea of replacing the multinomial models of an HMM by maximum entropy models is not new and has been applied before in different domains (Chen, 2003), as well as in POS induction (Berg-Kirkpatrick et al., 2010; Graça, 2010). A key advantage of this representation is that it allows for a much tighter control over the expressiveness of the model. For many languages it is helpful to exclude word identity features for rare words in order to constrain the model and force generalization across words with similar features. Unlike mapping all rare words to the 'unk' token in the multinomial setting, the maxent model still captures some information about the word through the other features. Moreover, we can reduce the number of parameters even further by using lowercase word identities while still keeping the case information by using a case feature. Table 1 shows the number of features we used for different corpora. Note that the reduced feature set has an order of magnitude fewer parameters than the multinomial model.

## 3. Learning

In Section 5 we describe experiments comparing the HMM model to the ME model under three learning scenarios: maximum likelihood training using the EM algorithm (Dempster, Laird, & Rubin, 1977) for both HMM and HMM+ME, gradient-based likelihood optimization for the HMM+ME model, and PR with sparsity constraints (Graça et al., 2009) for both HMM and HMM+ME. This section describes all three learning algorithms.

In the following, we denote the whole corpus, a list of sentences, by $\mathbf{X} = (\mathbf{x}^1, \mathbf{x}^2, \dots, \mathbf{x}^N)$ and the corresponding tag sequences by $\mathbf{Y} = (\mathbf{y}^1, \mathbf{y}^2, \dots, \mathbf{y}^N)$.

### 3.1 Maximum Likelihood with EM

Standard HMM training seeks model parameters $\theta$ that maximize the log-likelihood of the observed data:

$$\textbf{Log-Likelihood:} \ \ \mathcal{L}(\theta) = \log \sum_{\mathbf{Y}} p_\theta(\mathbf{X}, \mathbf{Y}) \tag{3}$$

where $\mathbf{X}$ is the whole corpus. Since the model assumes independence between sentences ,

$$\log \sum_{\mathbf{Y}} p_\theta(\mathbf{X}, \mathbf{Y}) = \sum_{n=1}^{N} \log \sum_{\mathbf{y}^n} p_\theta(\mathbf{x}^n, \mathbf{y}^n), \tag{4}$$

but we use the corpus notation for consistency with Section 3.3. Because of the latent variables $\mathbf{Y}$, the log-likelihood function for the HMM model is not convex in the model parameters, and the model is fitted using the EM algorithm. EM maximizes $\mathcal{L}(\theta)$ via block-coordinate ascent on a lower bound $F(q, \theta)$ using an auxiliary distribution over the latent variables $q(\mathbf{Y})$ (Neal & Hinton, 1998). By Jensen's inequality, we define a lower-bound $F(q, \theta)$ as:

$$\mathcal{L}(\theta) = \log \sum_{\mathbf{Y}} q(\mathbf{Y}) \frac{p_\theta(\mathbf{X}, \mathbf{Y})}{q(\mathbf{Y})} \geq \sum_{\mathbf{Y}} q(\mathbf{Y}) \log \frac{p_\theta(\mathbf{X}, \mathbf{Y})}{q(\mathbf{Y})} = F(q, \theta). \tag{5}$$

We can rewrite $F(q, \theta)$ as:

$$\begin{aligned} F(q, \theta) &= \sum_{\mathbf{Y}} q(\mathbf{Y}) \log(p_\theta(\mathbf{X}) p_\theta(\mathbf{Y}|\mathbf{X})) - \sum_{\mathbf{Y}} q(\mathbf{Y}) \log q(\mathbf{Y}) \tag{6} \\ &= \mathcal{L}(\theta) - \sum_{\mathbf{Y}} q(\mathbf{Y}) \log \frac{q(\mathbf{Y})}{p_\theta(\mathbf{Y}|\mathbf{X})} \tag{7} \\ &= \mathcal{L}(\theta) - \mathbf{KL}(q(\mathbf{Y})||p_\theta(\mathbf{Y}|\mathbf{X})). \tag{8} \end{aligned}$$

Using this interpretation, we can view EM as performing coordinate ascent on $F(q, \theta)$. Starting from an initial parameter estimate $\theta^0$, the algorithm iterates two block-coordinate ascent steps until a convergence criterion is reached:

$$\mathbf{E} : q^{t+1} = \arg\max_q F(q, \theta^t) = \arg\min_q \mathbf{KL}(q(\mathbf{Y}) \parallel p_{\theta^t}(\mathbf{Y} \mid \mathbf{X})) \tag{9}$$

$$\mathbf{M} : \theta^{t+1} = \arg\max_\theta F(q^{t+1}, \theta) = \arg\max_\theta \mathbf{E}_{q^{t+1}} \left[ \log p_\theta(\mathbf{X}, \mathbf{Y}) \right] \tag{10}$$

The E-step corresponds to maximizing Eq. 8 with respect to $q$ and the M-step corresponds to maximizing Eq. 6 with respect to $\theta$. The EM algorithm is guaranteed to converge to a local maximum of $\mathcal{L}(\theta)$ under mild conditions (Neal & Hinton, 1998). For an HMM POS tagger, the E-Step computes the posteriors $p_{\theta^t}(\mathbf{y}|\mathbf{x})$ over the latent variables (POS tags) given the observed variables (words) and current parameters $\theta^t$ for each sentence. This is accomplished by the forward-backward algorithm for HMMs. The EM algorithm together with the forward-backward algorithm for HMMs is usually referred to as the Baum–Welch algorithm (Baum, Petrie, Soules, & Weiss, 1970).

The M step uses $q^{t+1}$ ($q_n^{t+1}$ are the posteriors for a given sentence) to "fill in" the values of tags $\mathbf{Y}$ and estimate parameters $\theta^{t+1}$. Since the HMM model is locally normalized and the features used only depend on the tag and word identities and not on the particular position where they occur, the optimization decouples in the following way:

$$\mathbf{E}_{q^{t+1}}[\log p_\theta(\mathbf{X}, \mathbf{Y})] \quad = \quad \sum_{n=1}^{N} \mathbf{E}_{q_n^{t+1}}[\log \prod_{i=1}^{l_n} p_t(y_i^n \mid y_{i-1}^n) p_o(x_i^n \mid y_i^n)] \tag{11}$$

$$= \quad \sum_{n=1}^{N} \sum_{i=1}^{l_n} \left( \mathbf{E}_{q_n^{t+1}} \log p_t(y_i^n \mid y_{i-1}^n) + \mathbf{E}_{q_n^{t+1}} \log p_o(x_i^n \mid y_i^n) \right) \tag{12}$$

For the multinomial emission model, this optimization is particularly easy and simply involves normalizing (expected) counts for each parameter. For the maximum-entropy emission model parameterized as in Equation 2, there is no closed form solution so we need to solve an unconstrained optimization problem. For each possible hidden tag value $y$ we have to solve two problems: estimate the emission probabilities $p_o(x|y)$ and estimate the transition probabilities $p_t(y'|y)$, where the gradient for each one of those is given by

$$\frac{\partial \mathbf{E}_{q^{t+1}}[\log p_\theta(\mathbf{X}, \mathbf{Y})]}{\partial \theta} = \mathbf{E}_{q^{t+1}} \left[ \mathbf{f}(\mathbf{X}, \mathbf{Y}) - \mathbf{E}_{p_\theta(\mathbf{X'}|\mathbf{Y})}[\mathbf{f}(\mathbf{X'}, \mathbf{Y})] \right], \tag{13}$$

which is similar to the gradient in supervised ME models, except for the expectation over all $\mathbf{Y}$ under $q^{t+1}(\mathbf{Y})$ instead of observed $\mathbf{Y}$. The optimization is done using L-BFGS with Wolfe's rule line search (Nocedal & Wright, 1999).

### 3.2 Maximum Likelihood with Direct Gradient

While likelihood is traditionally optimized with EM, Berg-Kirkpatrick et al. (2010) find that for the HMM with the maximum entropy emission model, higher likelihood and better accuracy can be achieved by with a gradient-based likelihood-optimization method. They use L-BFGS in their experiments. The derivative of the likelihood is,

$$\frac{\partial \mathcal{L}(\theta)}{\partial \theta} \quad = \quad \frac{\partial}{\partial \theta} \log p_\theta(\mathbf{X}) = \frac{1}{p_\theta(\mathbf{X})} \frac{\partial}{\partial \theta} p_\theta(\mathbf{X}) = \frac{1}{p_\theta(\mathbf{X})} \frac{\partial}{\partial \theta} \sum_{\mathbf{Y}} p_\theta(\mathbf{X}, \mathbf{Y}) \tag{14}$$

$$= \quad \sum_{\mathbf{Y}} \frac{1}{p_\theta(\mathbf{X})} \frac{\partial}{\partial \theta} p_\theta(\mathbf{X}, \mathbf{Y}) = \sum_{\mathbf{Y}} \frac{p_\theta(\mathbf{X}, \mathbf{Y})}{p_\theta(\mathbf{X})} \frac{\partial}{\partial \theta} \log p_\theta(\mathbf{X}, \mathbf{Y}) \tag{15}$$

$$= \quad \sum_{\mathbf{Y}} p_\theta(\mathbf{Y}|\mathbf{X}) \frac{\partial}{\partial \theta} \log p_\theta(\mathbf{X}, \mathbf{Y}), \tag{16}$$

which is exactly the same as the derivative of the M-Step. Here in Equation 14 we apply the chain rule to take the derivative of $\log p_\theta(\mathbf{X})$, while in Equation 15 we apply the chain rule in the reverse direction. The biggest difference between the EM procedure and direct gradient is that for EM we fix the counts on the E-Step and optimize the ME model using those counts. When directly optimizing the likelihood we need to recompute the counts for each parameter setting, which can be expensive. Appendix A gives a more detailed discussion of both methods.

### 3.3 Controlling Tag Ambiguity with PR

One problem with unsupervised HMM POS tagging is that the maximum likelihood objective may encourage tag distributions that allow many different tags for a word in a given context. We do not find that in actual text with linguist-designed tags, because tags are designed to be informative about the word's grammatical role. In the following paragraphs we describe a measure of tag ambiguity proposed by Graça et al. (2009) that we will attempt to control. It is easier to understand this measure with hard tag assignments, so we start with that and thene extend the discussion to distributions over tags.

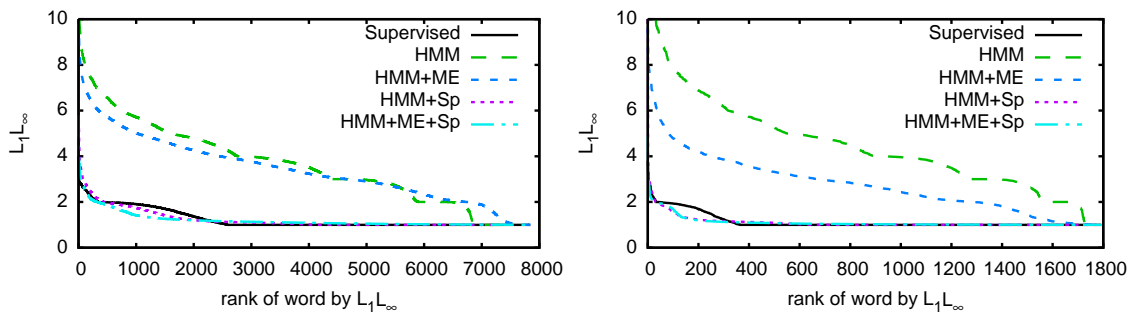

Figure 1: Our ambiguity measure ($\ell_1/\ell_\infty$) for each word type in two corpora for a supervised model, EM training (HMM, HMM+ME), and when we train with an ambiguity penalty as described in Section 3.3 (HMM+Sp, HMM+ME+Sp). Left:En, Right:Pt.

Consider a word such as "stock". Intuitively, we would like all occurrences of "stock" to be tagged with a small subset of all possible tags (noun and verb, in this case). For a hard assignment of tags to the entire corpus, $\mathbf{Y}$, we could count how many different tags are used in $\mathbf{Y}$ for occurrences of the word "stock."

If instead of a single tagging of the corpus, we have a distribution $q(\mathbf{Y})$ over assignments, we need to generalize this ambiguity measure. Instead of asking was a particular tag ever used for the word "stock", we would ask what is the maximum probability with which a particular tag was used for the word "stock". Then instead of counting the number of tags, we would sum these probabilities.

As motivation, Figure 1 shows the distribution of tag ambiguity across words for two corpora. As we see from Figure 1, when we train using the EM procedure described in Section 3.1, the HMM and ME models grossly overestimates the tag ambiguity of almost all words. However when the same models are trained using PR to penalize the tag ambiguity, both models (HMM+Sp, HMM+ME+Sp) achieve a tag ambiguity closer to the truth.

More formally, Graça et al. (2009) define this measure in terms of constraint features $\phi(\mathbf{X}, \mathbf{Y})$. Constraint feature $\phi_{wvj}(\mathbf{X}, \mathbf{Y})$ takes on value 1 if the $j^{th}$ occurrence of word type $w$ in $\mathbf{X}$ is assigned to tag $v$ in the tag assignment $\mathbf{Y}$. Consequently, the probability that the $j^{th}$ occurrence of word $w$ has tag $v$ under the label distribution $q(\mathbf{Y})$ is $\mathbf{E}_q[\phi_{wvj}(\mathbf{X}, \mathbf{Y})]$. The ambiguity measurement for word type $w$ becomes:

**Ambiguity Penalty for word type $w$:** $\qquad \sum_v \max_j \mathbf{E}_{q(\mathbf{Y})} \left[ \phi_{wvj}(\mathbf{X}, \mathbf{Y}) \right].$ (17)

This sum of maxima is also called the $\ell_1/\ell_\infty$ mixed norm. For brevity we use the norm notation $||\mathbf{E}_q[\phi_w]||_{1/\infty}$. For computational reasons, we do not add a penalty term based on the ambiguity of the model distribution $p_\theta(\mathbf{Y}|\mathbf{X})$, but instead introduce an auxiliary distribution $q(\mathbf{Y})$ which

must be close to $p_\theta$ but also must have low ambiguity. Our modified objective becomes

$$\max_{\theta,q} \ \mathcal{L}(\theta) - \mathbf{KL}(q(\mathbf{Y})||p_\theta(\mathbf{Y}|\mathbf{X})) - \sigma \sum_w ||\mathbf{E}_q[\phi_w(\mathbf{X},\mathbf{Y})]||_{1/\infty}.$$ (18)

Graça et al. (2009) optimize this objective using an algorithm very similar to EM. The added complexity of implementing their algorithm lies only in computing the Kullback-Leibler projection in a modified E-Step. However, this computation involves choosing a distribution over exponentially many objects (label assignments). Luckily, Graça et al. (2009) show that the dual formulation for the E-Step is more manageable. This is given by:

$$\max_{\boldsymbol{\lambda}\geq 0} \ -\log\left(\sum_{\mathbf{Y}} p_\theta(\mathbf{Y}|\mathbf{X})\exp(-\boldsymbol{\lambda}\cdot\phi(\mathbf{X},\mathbf{Y}))\right) \quad \text{s.t.} \quad \sum_j \lambda_{wvj}\leq\sigma$$ (19)

where $\boldsymbol{\lambda}$ is the vector of dual parameters $\lambda_{wvj}$, one for each $\phi_{wvj}$. The projected distribution is then given by: $q(\mathbf{Y})\propto p_\theta(\mathbf{Y}|\mathbf{X})\exp\left(-\boldsymbol{\lambda}\cdot\phi(\mathbf{X},\mathbf{Y})\right)$. Note that when $p_\theta$ is given by an HMM, $q$ for each sentence can be expressed as

$$q(\mathbf{y}^n)\propto\prod_{i=1}^{I} p_t(y_i^n\mid y_{i-1}^n)q_o(x_i^n\mid y_i^n),$$ (20)

where $q_o(x_i|y_i)=p_o(x_i|y_i)\exp(-\lambda_{x_i y_i j})$ act as modified (unnormalized) emission probabilities. The objective of Equation 19 is just the negative sum of the log probabilities of all the sentences under $q$ plus a constant. We can compute this by running forward-backward on the corpus, similar to the E-Step in normal EM. The gradient of the objective is also computed using the forward-backward algorithm. Note that the objective in Eq. 19 is concave with respect to $\lambda$ and can be optimized using a variety of methods. We perform the dual optimization by projected gradient, using the fast simplex projection algorithm for $\boldsymbol{\lambda}$ described by Bertsekas, Homer, Logan, and Patek (1995). In our experiments we found that taking a few projected gradient steps was not enough, and performing the optimization until convergence helps the results.

## 4. Related Work

POS tags place words into classes that share some commonalities as to what other (classes of) words they cooccur with. Therefore, it is natural to ask whether word clustering methods based on word context distributions might be able to recover the word classification inherent in a POS tag set. Several influential methods, most notably mutual-information clustering (Brown et al., 1992), have been used to cluster words according to how their immediately contiguous words are distributed. Although those methods were not explicitly designed for POS induction, the resulting clusters capture some syntactic information (see also Martin, Liermann, & Ney, 1998, for a different method with a similar objective). Clark (2003) refined the distributional clustering approach by adding morphological and word frequency information, to obtain clusters that more closely resemble POS tags.

Other forms of distributional clustering go beyond the immediate neighbors of a word to represent a whole vector of cooccurrences with the target word within a text window, and compare those vectors using some suitable metric, such as cosine similarity. However, these wider-range similarities have problems in capturing more local regularities. For instance, an adjective and a noun might look similar if the noun tends to be used in noun-noun compounds; similarly, two adjectives with different semantics or selectional preferences might be used with different contexts. Moreover, this problem is aggravated with data sparsity. As an example, infrequent adjectives that modify different nouns tend to have completely disjoint context vectors (but even frequent words like "a" and "an" might have completely different context vectors, since these articles are used in disjoint right contexts). To alleviate these problems, Schütze (1995) used frequency cutoffs, singular-value decomposition of co-occurrence matrices, and approximate co-clustering through two stages of SVD, with the clusters from the first stage used instead of individual words to provide vector representations for the second-stage clustering.

Lamar, Maron and Johnson (2010) have recently revised the two-stage SVD model of Schütze (1995) and achieve close to state-of-the-art performance. The revisions are relatively small, but touch several important aspects of the model: singular vectors are scaled by their singular values to preserve the geometry of the original space; latent descriptors are normalized to unit length; and cluster centroids are computed as a weighted average of their constituent vectors based on the word frequency, so that rare and common words are treated differently and centroids are initialized in a deterministic manner.

A final class of approaches – which include the work in this paper – uses a sequence model, such as an HMM, to represent the probabilistic dependencies between consecutive tags. In these approaches, each observation corresponds to a particular word and each hidden state corresponds to a cluster. However, as noted by Clark (2003) and Johnson (2007), using maximum likelihood training for such models does not achieve good results: maximum likelihood training tends to result in very ambiguous distributions for common words, in contradiction with the rather sparse word-tag distribution. Several approaches have been proposed to mitigate this problem. Freitag (2004) clusters the most frequent words using a distributional approach and co-clustering. To cluster the remaining (infrequent) words, the author trains a second-order HMM where the emission probabilities for the frequent words are fixed to the clusters found earlier and emission probabilities for the remaining words are uniform.

Several studies propose using Bayesian inference with an improper Dirichlet prior to favor sparse model parameters and hence indirectly reduce tag ambiguity (Johnson, 2007; Gao & Johnson, 2008; Goldwater & Griffiths, 2007). This was further refined by Moon, Erk, and Baldridge (2010) by representing explicitly the different ambiguity patterns of function and content words. Lee, Haghighi, and Barzilay (2010) take a more direct approach to reducing tag ambiguity by explicitly modeling the set of possible tags for each word type. Their model first generates a tag dictionary that assigns mass to only one tag for each word type to reflect lexicon sparsity. This dictionary is then used to constrain a Dirichlet prior from which the emission probabilities are drawn by only having support for word-tag pairs in the dictionary. Then a token-level HMM using those emission parameters and transition parameters draw from a symmetric Dirichlet prior are used for tagging the entire corpus. The authors also show improvements by using morphological features when creating the dictionary. Their system achieves state-of-art results for several languages. It should be noted that a common issue with the above sparsity-inducing approaches is that sparsity is imposed at the parameter level, the probability of word given tag, while the desired sparsity is at the posterior level, the probability of tag given word. Graça et al. (2009) use the PR framework to penalize ambiguous posteriors distributions of words given tokens, which achieves better results than the Bayesian sparsifying Dirichlet priors.

Most recently, Berg-Kirkpatrick et al. (2010) and Graça (2010) proposed replacing the multinomial distributions of the HMM by maximum entropy (ME) distributions. This allows the use of features to capture morphological information, and achieve very promising results. Berg-Kirkpatrick et al. (2010) also find that optimizing the likelihood with L-BFGS rather than EM leads to substantial improvements, which we show not to be the case beyond English.

We also note briefly POS induction methods that rely on a prior tag dictionary indicating for each word type what POS tags it can have. The POS induction task is then, for each word token in the corpus, to disambiguate between the possible POS tags, as described by Merialdo (1994). Unfortunately, the availability of a large manually-constructed tag dictionary is unrealistic and much of the later work tries to reduce the required dictionary size in different ways, by generalizing from a small dictionary with only a handful of entries (Smith & Eisner, 2005; Haghighi & Klein, 2006; Toutanova & Johnson, 2007; Goldwater & Griffiths, 2007). However, although this approach greatly simplifies the problem – most words can only have one tag and, furthermore, the cluster-tag mappings are predetermined, thus removing an extra level of ambiguity – the accuracy of such methods is still significantly behind supervised methods. To address the remaining ambiguity by imposing additional sparsity, Ravi and Knight (2009) minimize the number of possible tag-tag transitions in the HMM via a integer program. Finally, Snyder, Naseem, Eisenstein, and Barzilay (2008) jointly train a POS induction system over parallel corpora in several languages, exploiting the fact that different languages present different ambiguities.

## 5. Experiments

In this section we present encouraging results validating the proposed method in six different testing scenarios according to different metrics. The highlights are:

- A maximum-entropy emission model with a Markov transition model trained with the ambiguity penalty improves over the regular HMM in all cases with an average improvement of 10.4% (according to the 1-Many metric).

- When compared against a broad range of recent POS induction systems, our method produces the best results for all languages except English. Furthermore, the method seems less sensitive to particular test conditions than previous methods.

- The induced clusters are useful features in training supervised POS taggers, improving test accuracy as much or more than the clusters learned by competing methods.

### 5.1 Corpora

In our experiments we test several POS induction methods on five languages with the help of manually POS-tagged corpora for those languages. Table 1 summarizes characteristics of the test corpora: the Wall Street Journal portion of the Penn Treebank (Marcus et al., 1993) (we consider both the 17-tag version of Smith & Eisner, 2005 (En17) and the 45-tag version (En45)); the Bosque subset of the Portuguese Floresta Sinta(c)tica Treebank (Afonso, Bick, Haber, & Santos, 2002) (Pt); the Bulgarian BulTreeBank (Simov et al., 2002) (Bg) (with only the 12 coarse tags); the Spanish corpus from the Cast3LB treebank (Civit & Martí, 2004) (Es); and the Danish Dependency Treebank (DDT) (Kromann, Matthias T., 2003) (Dk).

| | 1 | 2 | 3 | 4 | 5 | 6 | 7 | 8 | 9 |
|---|---|---|---|---|---|---|---|---|---|
| | Sentences | Types | LUnk | Tokens | Tags | Avg. $\ell_1/\ell_\infty$ | Total $\ell_1/\ell_\infty$ | $|w|^1$ | $|w|^2$ |
| En | 49208 | 49206 | 49.28% | 1173766 | 17 (45) | 1.08 (1.11) | 6523.6 (6746.2) | 54334 | 7856 |
| Pt | 9359 | 29489 | 37.83% | 212545 | 22 | 1.02 | 1144.6 | 33293 | 2114 |
| Bg | 14187 | 34928 | 39.26% | 214930 | 12 | 1.02 | 3252.9 | 38633 | 2287 |
| Es | 3512 | 16858 | 37.73% | 95028 | 47 | 1.05 | 818.9 | 18962 | 951 |
| Dk | 5620 | 19400 | 26.30% | 65057 | 30 | 1.05 | 795.8 | 21678 | 969 |

Table 1: Corpus statistics. The third column shows the percentage of word types after lower-casing and eliminating word types occurring only once. The sixth and seventh columns show information about the word ambiguity in each corpus on average and in totality (corresponding to the penalty in Equation 17). The eighth and ninth columns show the number of parameters for the different feature sets, as described in Section 5.3.

## 5.2 Experimental Setup

We compare our work with two kinds of methods: those that induce a single cluster for each word type (type-level tagging), and those that allow different tags on different occurrences of a word type (token-level tagging). For type-level tagging, we use two standard baselines, Brown and Clark, as described by Brown et al. (1992)[1] and Clark (2003)[2]. Following Headden, McClosky, and Charniak (2008), we trained the Clark system with both 5 and 10 hidden states for the letter HMM and ran it for 10 iterations; the Brown system was run according with the instructions accompanying the code. We also ran the recently proposed LDC system (Lamar, Maron, & Bienenstock, 2010)[3], with the configuration described in their paper for PTB45 and PTB17, and the PTB17 configuration for the other corpora. It should be noted that we did not carry out our experiments with the SVD2 system (Lamar, Maron and Johnson, 2010), since SVD2 is superseded by LDC according to its authors.

For token-level tagging, we experimented with the feature-rich HMM as presented by Berg-Kirkpatrick et al. (2010), trained both using EM training (BK+EM) and direct gradient (BK+DG), using the configuration provided by the authors[4]. We report results from the type-level HMM (TLHMM) (Lee et al., 2010) when applicable, since we were not able to run that system. Moreover, we compared those systems against our own implementation of various HMM-based approaches: the HMM with a multinomial emission probabilities (Section 2.1), the HMM with maximum-entropy emission probabilities (Section 2.2) trained with EM (HMM+ME), trained by direct gradient (HMM+ME+DG), and trained using PR with the ambiguity penalty, as described in Section 3.3 (HMM+Sp for multinomial emissions, and HMM+ME+Sp for maximum-entropy emissions). In addition, we also compared to a multinomial HMM with a sparsifying Dirichlet prior on the parameters (HMM+VB) trained using variational Bayes (Johnson, 2007).

Following standard practice, for the multinomial HMMs that do not use morphological information, we lowercase the corpora and replace unique words by a special *unknown* token, as this improves the multinomial HMM results by decreasing the number of parameters and eliminating very rare words (mostly nouns). Since the maximum-entropy emission models have access to morphological features, these preprocessing steps do not improve performance and we do not perform them in that case.

---

1. Implementation: `http://www.cs.berkeley.edu/~pliang/software/brown-cluster-1.2.zip`

2. Implementation: `http://www.cs.rhul.ac.uk/home/alexc/pos2.tar.gz`

3. Implementation provided by Lamar, Maron and Bienenstock (2010).

4. Implementation provided by Berg-Kirkpatrick et al. (2010).

At the start of EM, we randomly initialize all of our implementations of HMM-based models from the same posteriors, obtained by running the E-step of the HMM model with a set of random parameters: close to uniform with random uniform "jitter" of 0.01. This means that for each random seed, the initialization is identical for all models.

For EM and variational Bayes training, we train the model for 200 iterations, since we found that typically most models tend to converge by iteration 100. For the HMM+VB model we fix the transition prior[5] $\alpha$ to 0.001 and test an emission prior $\alpha$ equal to 0.1 and 0.001, corresponding to the best values reported by Johnson (2007).

For PR training, we initialize with 30 EM iterations and then run for 170 iterations of PR, following Graça et al. (2009). We used the results that worked best for English (En17) (Graça et al., 2009), regularizing only words that occur at least 10 times, with $\sigma = 32$, and use the same configuration for all the other scnenarios. This setting was not specifically tuned for the test languages, and might not be optimal for every language. Setting such parameters in an unsupervised manner is a difficult task and we do not address it here (Graça, 2010 discusses more experiments with different values of those parameters).

We obtain hard assignments using posterior decoding, where for each position we pick the label with highest posterior probability, since this showed small but consistent improvements over Viterbi decoding. For all experiments that required random initialization of the parameters we report the average of 5 random seeds.

All experiments were run using the number of true tags as the number of clusters, with results obtained in the test set portion of each corpus. We evaluate all systems using four common metrics for POS induction: **1-Many** mapping, **1-1** mapping (Haghighi & Klein, 2006), variation of information (**VI**) (Meilă, 2007), and validity measure (**V**) (Rosenberg & Hirschberg, 2007). These metrics are described in detail in Appendix B.

### 5.3 HMM+ME+Sp Performance

This section compares the gains from using a feature-rich representation with those from the ambiguity penalty, as described in Section 3.3. Experiments show that having a feature-rich representation always improves performance, and that having an ambiguity penalty also always improves performance. Then, we will see that the improvements from the two methods combine additively, suggesting that they address independent aspects of POS induction.

We use two different feature sets: the *large feature set* is that of Berg-Kirkpatrick et al. (2010), while the *reduced feature set* was described by Graça (2010). We apply count-based feature selection to both the identity and suffix features. Specifically, we only add identity features for words occurring at least 10 times and suffix features for words occurring at least 20 times. We also add a punctuation feature. In what follows, we refer to the large feature set as feature set 1 and the reduced feature set as 2. The total number of features for each model and language is given in Table 1. The results of these experiments are summarized in Table 2.

Table 2 shows the results for 10 training methods across six corpora and four evaluation metrics, resulting in 240 experimental conditions. To simplify the discussion, we focus on the 1-Many metric (top left tab of Table 2), and just observe that the conclusions hold for the other three evaluation metrics also. From Table 2 we can conclude the following:

- Adding a penalty for high word-tag ambiguity improves the performance of the multinomial HMM. The multinomial HMM trained with EM (line 1 in Table 2) is always worse than the multinomial HMM trained with PR and an ambiguity penalty, by 6.5% on average (line 2 in Table 2).

- The feature-rich maximum entropy HMMs (lines 3-6 in Table 2) almost always perform better than the multinomial HMM. This is true for both feature sets and both regularization strengths used, with an average increase of 6.4%. The exceptions are possibly due to suboptimal regularization.

- Adding a penalty for high word-tag ambiguity to the maximum-entropy HMM improves performance. In almost all cases, comparing lines 3-6 to lines 7-10 in Table 2, the sparsity constraints improve performance (average improvement of 1.6%). The combined system al-most always outperforms the multinomial HMM trained using the ambiguity penalty with an average improvement of 1.6%. For every corpus the best performance is achieved by the model with an ambiguity penalty and maximum-entropy emission probabilities.

- For every language except English with 17 tags and a particular feature configuration, reducing the feature set by excluding rare features improves performance on average by 2.3% (lines 5-6 are better than lines 3-4 in Table 2).

- Regularizing the maximum-entropy model is more important when there are many features and when we do not have a word-tag ambiguity penalty. Lines 3-4 of Table 2 have the maximum-entropy HMM with many features, and we see that having a tight parameter prior almost always out-performs having a looser prior. By contrast, looking at lines 9-10 of Table 2 we see that when we have an ambiguity penalty and fewer features a looser prior is almost always better than a tighter parameter prior. This was observed also by Graça (2010).

---

5. The transition prior does not significantly affect the results, and we do not report results with different values.

| | | 1-Many | | | | | | 1-1 | | | | | |
|---|---|---|---|---|---|---|---|---|---|---|---|---|---|
| | | En45 | En17 | PT | BG | DK | ES | PTB45 | PTB17 | PT | BG | DK | ES |
| 1 | HMM | 62.4 | 65.6 | 64.9 | 58.9 | 60.0 | 60.2 | 42.5 | 43.5 | 42.2 | 40.6 | 37.4 | 30.6 |
| 2 | HMM+Sp | 67.5 | 70.3 | 71.2 | 65.0 | 67.5 | 69.0 | 46.1 | 52.5 | 47.7 | 46.3 | 40.0 | 35.3 |
| 3 | HMM+ME$^1$ Prior 1 | 67.1 | 72.2 | 72.3 | 61.1 | 65.1 | 71.8 | 45.0 | 51.1 | 46.3 | 46.1 | 42.6 | 40.8 |
| 4 | HMM+ME$^1$ Prior 10 | 70.0 | 69.2 | 66.8 | 58.2 | 63.9 | 66.2 | 48.4 | 45.6 | 40.8 | 41.7 | 41.1 | 35.1 |
| 5 | HMM+ME$^2$ Prior 1 | 70.3 | 71.8 | 73.9 | 62.4 | 66.5 | 72.9 | 45.1 | 49.8 | 46.8 | 46.7 | 45.0 | 42.6 |
| 6 | HMM+ME$^2$ Prior 10 | 69.9 | 71.0 | 74.1 | 63.9 | 67.8 | 73.4 | 47.3 | 51.2 | 48.8 | 48.8 | 44.7 | 37.1 |
| 7 | HMM+ME+Sp$^1$ Prior 1 | 71.4 | 71.7 | 75.1 | 63.0 | 64.5 | 72.8 | 49.4 | 52.5 | 46.6 | 46.7 | 43.1 | 41.1 |
| 8 | HMM+ME+Sp$^1$ Prior 10 | 68.8 | 71.1 | 71.6 | 62.2 | 68.2 | 69.3 | 45.1 | 52.1 | 48.0 | 49.7 | 42.0 | 38.5 |
| 9 | HMM+ME+Sp$^2$ Prior 1 | **71.6** | **72.5** | 73.4 | 60.9 | 65.0 | 72.1 | **52.5** | **53.9** | 45.5 | 49.5 | 43.0 | 38.6 |
| 10 | HMM+ME+Sp$^2$ Prior 10 | 71.1 | 72.0 | **76.9** | **67.1** | **72.0** | **75.2** | 46.7 | 48.5 | **49.6** | **53.4** | **48.7** | 40.8 |

| | | VI | | | | | | V | | | | | |
|---|---|---|---|---|---|---|---|---|---|---|---|---|---|
| | | En45 | En17 | PT | BG | DK | ES | PTB45 | PTB17 | PT | BG | DK | ES |
| 1 | HMM | 4.22 | 3.75 | 3.90 | 4.04 | 4.55 | 4.89 | .558 | .479 | .490 | .383 | .432 | .474 |
| 2 | HMM+Sp | 3.64 | 3.20 | 3.27 | 3.49 | 3.85 | 3.85 | .616 | .549 | .573 | .467 | .518 | .581 |
| 3 | HMM+ME$^1$ Prior 1 | 3.77 | 3.11 | 3.21 | 3.46 | 3.77 | 3.56 | .606 | .564 | .583 | .460 | .519 | .608 |
| 4 | HMM+ME$^1$ Prior 10 | 3.31 | 3.38 | 3.66 | 3.83 | 4.13 | 4.11 | .649 | .530 | .527 | .406 | .482 | .553 |
| 5 | HMM+ME$^2$ Prior 1 | 3.59 | 3.12 | 3.05 | 3.46 | 3.71 | 3.46 | .626 | .559 | .600 | .460 | .528 | .617 |
| 6 | HMM+ME$^2$ Prior 10 | 3.28 | 3.24 | 3.12 | 3.44 | 3.74 | 3.58 | .652 | .546 | .596 | .471 | .530 | .610 |
| 7 | HMM+ME+Sp$^1$ Prior 1 | **3.20** | 3.09 | **3.00** | 3.38 | 3.80 | 3.49 | **.660** | .560 | .608 | .478 | .514 | .617 |
| 8 | HMM+ME+Sp$^1$ Prior 10 | 3.46 | 3.15 | 3.15 | 3.43 | 3.73 | 3.76 | .637 | .557 | .591 | .470 | .532 | .589 |
| 9 | HMM+ME+Sp$^2$ Prior 1 | 3.21 | **3.04** | 3.16 | 3.37 | 3.72 | 3.46 | .658 | **.567** | .591 | .473 | .523 | .616 |
| 10 | HMM+ME+Sp$^2$ Prior 10 | 3.41 | 3.25 | **2.86** | **3.12** | **3.35** | 3.34 | .644 | .541 | **.631** | **.519** | **.578** | **.636** |

Table 2: Results for different HMMs. HMM and HMM+Sp are HMMs with multinomial emission functions trained using EM and PR with sparsity constraints, respectively. HMM+ME and HMM+ME+Spare HMMs with a maximum entropy emission model trained using EM and PR with sparsity constraints. For the feature-rich models, superscript $^1$ represents the large feature set, and superscript $^2$ represents the reduced feature set. Prior 1 and 10 refers to the regularization strength for the ME emission model. Table entries are results averaged over 5 runs. **Bold** indicates best system overall.

It is very encouraging to see that the improvements of using a feature-rich model are additive with the effects of penalizing tag-ambiguity. This is especially surprising since we did not optimize the strength of the tag-ambiguity penalty for the maximum-entropy emission HMM, but rather used a value reported by Graça et al. (2009) to work for the multinomial emission HMM. Experiments reported by Graça (2010) show that tuning this parameter can further improve performance. Nevertheless, both methods regularize the objective in different ways and their interaction should be accounted for. It would be interesting to use $L_1$ regularization on the ME models, instead of $L_2^2$ regularization together with a feature count cutoff. This way the model could learn which features to discard, instead of requiring a predefined parameter that depends on the particular corpus characteristics.

As reported by Berg-Kirkpatrick et al. (2010), the way in which the objective is optimized can have a big impact on the overall results. However, due to the non-convex objective function it is unclear which optimization method works better and why. We briefly analyze this question in Appendix A and leave it as an open question for future work.

### 5.4 Error Analysis

Figure 2 shows the distribution of true tags and clusters for both the HMM model (left) and the HMM+ME+Sp model (right) on the En17 corpus. Each bar represents a cluster, labeled by the tag assigned to it after performing the 1-Many mapping. The colors represent the number of words with the corresponding true tag. To reduce clutter, true tags that were never used to label a cluster are grouped into "Others".

We observe that both models split common tags such as "nouns" into several hidden states. This splitting accounts for many of the errors in both models. By using 5 states for nouns instead of 7, HMM+ME+Sp is able to use more states for adjectives. Another improvement comes from a better grouping of prepositions. For example "to" is grouped with punctuation by the HMM while for HMM+ME+Sp it is correctly mapped to prepositions. Although this should be the correct behavior, it actually hurts, since the tagset has a special tag "TO" and all occurrences of the word "to" are incorrectly assigned, resulting in the loss of 2.2% accuracy. In contrast, HMM has a state mapped to the tag "TO" but the word "to" comprises only one fifth of that state. The most common error made by HMM+ME+Sp is to include the word "The" with the second noun induced tag in Figure 2 (Right). This induced tag contains mostly capitalized nouns and pronouns, which often precede nouns of other induced tags. We suspect that the capitalization feature is the cause of this error.

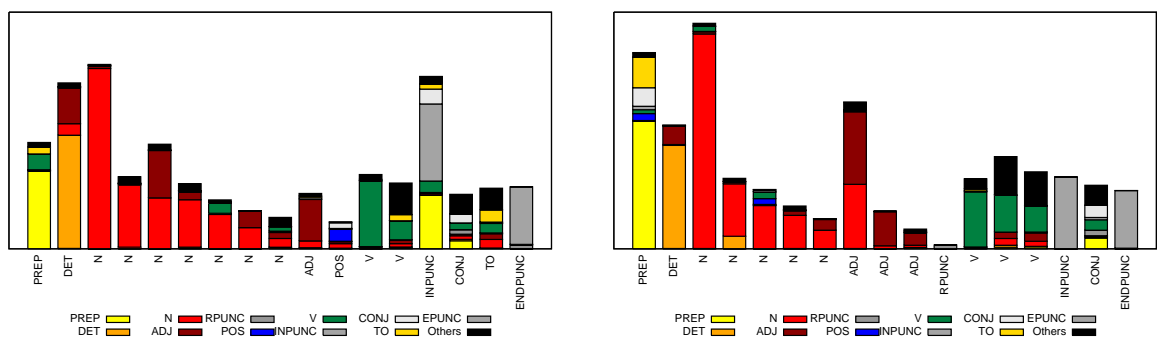

Figure 2: Induced tags by the HMM model (Left), and by the HMM+ME+Sp model (Right) on the En17 corpus. Each column represents a hidden state, and is labeled by its 1-Many mapping. Unused true tags are grouped into the cluster named "Others".

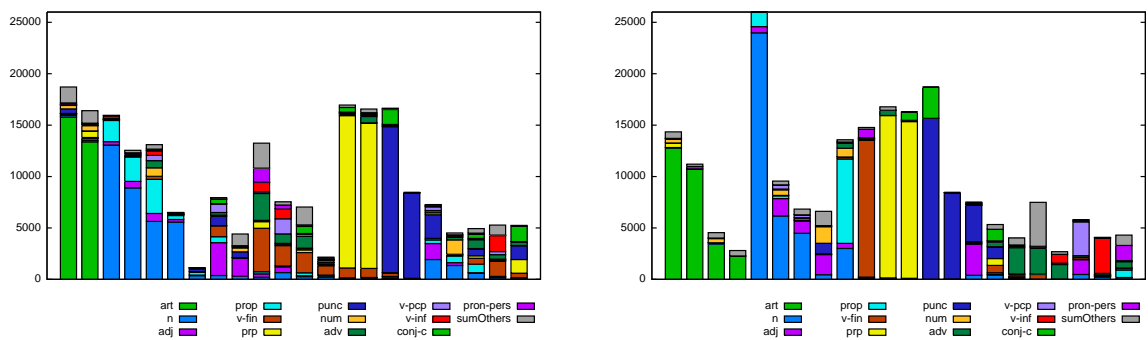

Figure 3: Induced tags by the HMM model (Left), and by the HMM+ME+Sp model (Right) on the Pt corpus. Each column represents a hidden state, and is labeled by its 1-Many mapping. Unused true tags are grouped into the cluster named "Others".

The better performance of feature-based models on Portuguese relative to English may be due to the ability of features to better represent the richer morphology of Portuguese. Figure 3 shows the induced clusters for Portuguese. The HMM+ME+Sp model improves over HMM for all tags except for adjectives. Both models have trouble distinguishing nouns from adjectives. The reduced accuracy for adjectives for HMM+ME+Sp is explained by the mapping of a single cluster containing most of the adjectives to adjectives by the HMM model and to nouns in the HMM+ME+Sp model. Removing the noun-adjective distinction, as suggested by Zhao and Marcus (2009), would increase performance of both models by about 6%. Another qualitative difference we observed was that the HMM+ME+Sp model used a single induced cluster for proper nouns rather than spreading them across different clusters.

## 5.5 State-of-the-Art Comparison

We now compare our best POS induction system (based on the settings in line 10 of Table 2), to other recent systems. Results are summarized in Table 3. As we have previously done with Table 2, we focus the discussion on the 1-Many evaluation metric, as results are qualitatively the same for the VI and V metrics, while the 1-1 metric shows more variance across languages.

| | | 1-Many | | | | | | 1-1 | | | | |
|---|---|---|---|---|---|---|---|---|---|---|---|---|
| | | En45 | En17 | PT | BG | DK | ES | PTB45 | PTB17 | PT | BG | DK | ES |
| 1 | BROWN | 68.7 | 68.7 | 69.6 | 63.2 | 69.6 | 69.7 | 53.3 | **56.6** | 43.8 | 47.9 | 44.3 | 41.2 |
| 2 | CLARK5 | 72.4 | 63.5 | 66.0 | 62.3 | 57.3 | 67.6 | 52.3 | 43.1 | 47.3 | 50.3 | 37.5 | 37.4 |
| 3 | CLARK10 | 72.5 | 63.2 | 67.1 | 57.0 | 58.2 | 70.1 | 52.3 | 44.2 | 48.1 | 45.2 | 37.5 | 40.0 |
| 4 | LDC | 67.5 | **74.7** | 69.2 | 61.1 | 60.9 | 67.9 | 48.6 | 50.0 | 42.6 | 50.1 | 35.2 | 38.8 |
| 5 | HMM | 62.4 | 65.6 | 64.9 | 58.9 | 60.0 | 60.2 | 42.5 | 43.5 | 42.2 | 40.6 | 37.4 | 30.6 |
| 6 | HMM+VB0.1 | 55.0 | 67.2 | 61.5 | 51.5 | 51.2 | 45.5 | 48.1 | 50.3 | 51.4 | 42.0 | 42.7 | 35.9 |
| 7 | HMM+VB0.001 | 58.6 | 67.7 | 63.2 | 53.5 | 56.6 | 55.9 | 44.1 | 51.4 | 45.1 | 38.3 | 38.1 | 34.4 |
| 8 | HMM+Sp | 67.5 | 70.3 | 71.2 | 65.0 | 67.5 | 69.0 | 46.1 | 52.5 | 47.7 | 46.3 | 40.0 | 35.3 |
| 9 | BK+EM | 69.1 | 72.1 | 72.3 | 64.3 | 62.8 | 72.0 | 48.3 | 54.4 | 45.5 | 50.6 | 41.5 | 37.2 |
| 10 | BK+DG | **75.8** | 67.9 | 72.5 | 56.1 | 60.6 | 73.7 | **54.5** | 47.9 | 42.9 | 38.8 | 41.5 | 40.4 |
| 11 | TLHMM | 62.2 | | 74.5 | | 61.2 | 68.9 | 50.9 | | **64.1** | | **52.1** | **58.3** |
| 12 | HMM+ME+Sp² Prior 10 | 71.1 | 72.0 | **76.9** | **67.1** | **72.0** | **75.2** | 46.7 | 48.5 | 49.6 | **53.4** | 48.7 | 40.8 |

| | | VI | | | | | | V | | | | |
|---|---|---|---|---|---|---|---|---|---|---|---|---|
| | | En45 | En17 | PT | BG | DK | ES | PTB45 | PTB17 | PT | BG | DK | ES |
| 1 | BROWN | 3.17 | 3.07 | 3.34 | 3.30 | 3.41 | 3.40 | .648 | .559 | .564 | .473 | .554 | .613 |
| 2 | CLARK5 | 3.23 | 3.45 | 3.38 | 3.30 | 3.97 | 3.76 | .660 | .498 | .545 | .475 | .490 | .588 |
| 3 | CLARK10 | 3.20 | 3.46 | 3.28 | 3.60 | 3.99 | 3.55 | .663 | .499 | .557 | .424 | .485 | .610 |
| 4 | LDC | 3.43 | **3.00** | 3.50 | 3.51 | 4.24 | 4.00 | .626 | **.585** | .546 | .450 | .474 | .569 |
| 5 | HMM | 4.22 | 3.75 | 3.90 | 4.04 | 4.55 | 4.89 | .558 | .479 | .490 | .383 | .432 | .474 |
| 6 | HMM+VB0.1 | 4.10 | 3.52 | 3.65 | 3.90 | 4.20 | 4.40 | .534 | .500 | .477 | .368 | .405 | .402 |
| 7 | HMM+VB0.001 | 4.38 | 3.55 | 3.97 | 4.07 | 4.41 | 4.69 | .535 | .501 | .471 | .368 | .437 | .474 |
| 8 | HMM+Sp | 3.64 | 3.20 | 3.27 | 3.49 | 3.85 | 3.85 | .616 | .549 | .573 | .467 | .518 | .581 |
| 9 | BK+EM | 3.31 | 3.04 | 3.19 | 3.30 | 3.90 | 4.15 | .645 | .568 | .582 | .479 | .477 | .596 |
| 10 | BK+DG | **3.01** | 3.21 | 3.29 | 3.89 | 4.15 | 3.56 | **.678** | .534 | .574 | .392 | .477 | .611 |
| 12 | HMM+ME+Sp² Prior 10 | 3.41 | 3.25 | **2.86** | **3.12** | **3.35** | **3.34** | .644 | .541 | **.631** | **.519** | **.578** | **.636** |

Table 3: Comparing HMM+ME+Sp² with several other POS induction systems. All results from models with random initialization that we run (systems: 2,3,5,6,7,8,9,10,12) represent an average over 5 runs. See Section 5.5 for details and discussion.

Lines 1-3 in Table 3 show clustering algorithms based on the information gain on various metrics. BROWN wins 5/6 times (in the scenarios with fewer clusters) over the CLARK system, despite the fact that CLARK uses morphology. Comparing lines 1-3 of Table 3 to line 4, we see that the LDC system is particularly strong for En17 where it achieves state-of-the-art results, but behaves worse than the BROWN system for every other corpus.

For HMMs with multinomial emissions (lines 5-8 of Table 3), both maximum likelihood training (HMM) and parameter sparsity (HMM+VB) perform worse than adding an ambiguity penalty (HMM+Sp). This holds for other evaluation metrics, with the exception of 1-1. This confirms previous results by Graça et al. (2009). Comparing the models in lines 5-8 to those in lines 1-3, we see that the best HMM (HMM+Sp) performs comparably with the best clustering (BROWN), with one model winning for 3 languages and the other for the remaining 3.

The feature rich HMMs (BK+EM and BK+DG) perform very well, achieving results that are better than HMM+Sp for 4 of 6 tests. Even though both optimize the same objective, they achieve different results on different corpora. We explore the training procedure in more detail in Appendix A, comparing also our implementation to that of Berg-Kirkpatrick et al. (2010). For brevity, Table 3 contains only the results from the implementation of Berg-Kirkpatrick et al. (2010). Our implementation produces comparable, but not quite identical results.

Lines 11-12 of Table 3 display the two methods that attempt to control tag ambiguity and have a feature-rich representation to capture morphological information. The results for TLHMM are taken from Lee et al. (2010), so we do not report results for the En17 and Bg corpora. Also, because we were not able to rerun the experiments for TLHMM, we were not able to compute the information-theoretic metrics. Consequently, the comparison for TLHMM is slightly less complete than for the other methods. Both TLHMM and HMM+Sp perform competitively or better than the other systems. This is not surprising since they have the ability to model morphological regularity while also penalizing high ambiguity. Comparing TLHMM with HMM+ME+Sp, we see that HMM+ME+Sp performs better on the 1-Many metric. In contrast, TLHMM performs better on 1-1. One possible explanation is that the underlying model in TLHMM is a Bayesian HMM with sparsifying Dirichlet priors. As noted by Graça et al. (2009), models trained in this way tend to have a cluster distribution that more closely resemble the true POS distribution (some clusters with lots of words and some with few words) which favors the 1-1 metric (a description of the particularity of the 1-1 metric is discussed in Appendix B).

To summarize, for all non-English languages and all metrics except 1-1, the HMM+ME+Sp system performs better than all the other systems. For English, BK+DG wins for the 45-tag corpus, while LDC wins for the 17-tag corpus. The HMM+ME+Sp system is fairly robust, performing well on all corpora and best on several of them, which allow us to conclude that it is not tuned to any particular corpus or evaluation metric.

The performance of HMM+ME+Sp is tightly related to the performance of the underlying HMM+ME system. In Appendix A we present a discussion about the performance of different optimization methods for HMM+ME. We compare our HMM+ME implementation to that of BK+EM and BK+DG and show that there are some significant differences in performance. However, its not clear by the results which one is better, and why it performs better in a given situation.

As mentioned by Clark (2003), morphological information is particularly useful for rare words. Table 4 compares different models' accuracy for words according to their frequency. We compare clustering models based on information gain with and without morphological information (BROWN,CLARK), a distributional information-based model (LDC), and the feature rich HMM with tag ambiguity control (HMM+ME+Sp). As expected we see that systems using morphology do better on rare words. Moreover these systems improve over almost all categories except very common words (words occurring more than 50 times). Comparing HMM+ME+Sp against CLARK, we see that even for the condition where CLARK overall works better (En45), it still performs worse for rare words than HMM+ME+Sp.

| | En45 | | | | En17 | | | |
|---|---|---|---|---|---|---|---|---|
| | BROWN | CLARK | LDC | HMM+ME+Sp$^2$ | BROWN | CLARK | LDC | HMM+ME+Sp$^2$ |
| $\leq 1$ | 50.07 | 49.48 | 50.65 | **70.12** | 29.42 | 60.62 | 53.39 | **75.79** |
| $\leq 5$ | 61.76 | 62.89 | 58.70 | **72.12** | 43.38 | 69.30 | 64.03 | **76.50** |
| $\leq 10$ | 64.32 | 66.53 | 61.78 | **70.59** | 50.94 | 71.13 | 67.35 | **74.29** |
| $\leq 50$ | 67.13 | 67.62 | 64.28 | **70.80** | 59.16 | 71.50 | 68.02 | **75.31** |
| $> 50$ | 68.94 | **73.87** | 68.04 | 71.49 | 72.14 | 62.04 | **77.14** | 71.40 |
| | PT | | | | BG | | | |
| | BROWN | CLARK | LDC | HMM+ME+Sp$^2$ | BROWN | CLARK | LDC | HMM+ME+Sp$^2$ |
| $\leq 1$ | 31.44 | 44.53 | 19.19 | **63.51** | 32.55 | 52.25 | 40.61 | **68.17** |
| $\leq 5$ | 48.06 | 53.30 | 36.67 | **68.60** | 48.18 | 65.00 | 53.58 | **73.54** |
| $\leq 10$ | 63.14 | 64.33 | 54.16 | **72.35** | 56.59 | 69.51 | 60.82 | **71.53** |
| $\leq 50$ | 66.30 | 67.52 | 62.56 | **71.32** | 60.15 | **68.95** | 62.06 | 68.59 |
| $> 50$ | 80.61 | 74.68 | **84.58** | 79.52 | **74.01** | 61.89 | 67.02 | 65.89 |
| | ES | | | | DK | | | |
| | BROWN | CLARK | LDC | HMM+ME+Sp$^2$ | BROWN | CLARK | LDC | HMM+ME+Sp$^2$ |
| $\leq 1$ | 41.04 | 49.93 | 38.75 | **68.65** | 37.71 | 43.58 | 35.65 | **64.41** |
| $\leq 5$ | 61.75 | 64.03 | 51.94 | **72.05** | 49.54 | 48.62 | 40.17 | **66.98** |
| $\leq 10$ | 72.10 | 69.76 | 61.69 | **73.35** | 58.90 | 51.92 | 47.96 | **65.21** |
| $\leq 50$ | **64.44** | 57.67 | 56.74 | 59.94 | 60.42 | 55.87 | 46.12 | **61.83** |
| $> 50$ | **86.43** | 73.69 | 81.94 | 82.11 | **82.82** | 60.88 | 77.91 | 73.26 |

Table 4: 1-Many accuracy by word frequency for different corpora.

## 5.6 Using the Clusters

As a further comparison of the different POS induction methods, we experiment with a simple semisupervised scheme where we use the learned clusters as features in a supervised POS tagger. The basic supervised model has the same features as the HMM+ME model, except that we use all word identities and suffixes regardless of frequency. We trained the supervised model using averaged perceptron for a number of iterations chosen as follows: split the training set into 20% for development and 80% for training and pick the number of iterations $\nu$ to optimize accuracy on the development set. Finally, trained on the full training set using $\nu$ iterations and report results on a 500 sentence test set.

We augmented the standard features with the learned hidden state for the current token, for each unsupervised method (BROWN,CLARK,LDC, HMM+ME+Sp). Figure 4 shows the average accuracy of the supervised model as we varied the type of unsupervised features. The average is taken over 10 random samples for the training set at each training set size. We can see from Figure 4 that using sem-supervised features from any of the models improves performance even if we have 500 labeled sentences. Moreover, we see that HMM+ME+Sp either performs as well or better than the other models.

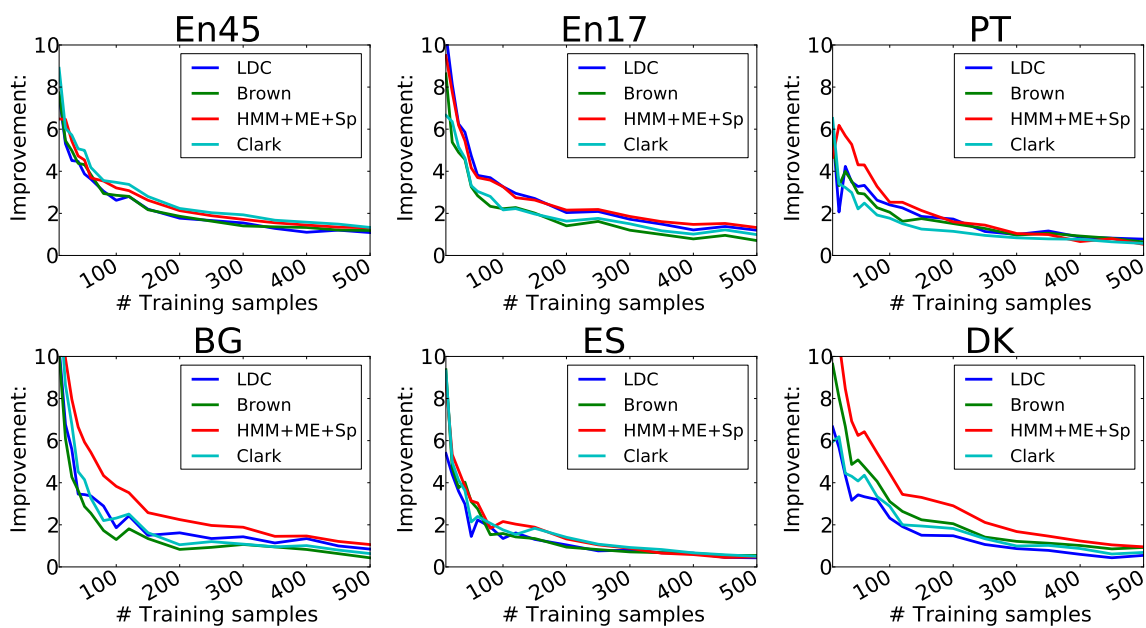

Figure 4: Error reduction from using the induced clusters as features on a semi-supervised model as a function of labeled data size. Top Left: En45. Top Middle: En17. Top Right: PT. Bottom Left: BG. Bottom Middle: ES. Bottom Right: DK.

## 6. Conclusion

In this work we investigated the task of fully unsupervised POS induction in five different languages. We identified and proposed solutions for three major problems of the simple hidden Markov model that has been used extensively for this task: i) treating words atomically, ignoring orthographic and morphological information – which we addressed by replacing multinomial word distributions by small maximum-entropy models; ii) an excessive number of parameters that allows models to fit irrelevant correlations – which we adressed by discarding parameters with small support in the corpus; iii) a training regime (maximum likelihood) that allows very high word ambiguity – which we addressed by training using the PR framework with a word ambiguity penalty. We show that all these solutions improve the model performance and that the improvements are additive. Comparing against the regular HMM we achieve an impressive improvement of 10.4% on average.

We also compared our system against the main competing systems and show that our approach performs better in every language except English. Moreover, our approach performs well across languages and learning conditions, even when hyperparameters are not tuned to the conditions. When the induced clusters are used as features in a semi-supervised POS tagger trained with a small amount of supervised data, we show significant improvements. Moreover, the clusters induced by our system always perform as well as or better than the clusters produced by other systems.

**Acknowledgments**

João V. Graça was supported by a fellowship from Fundação para a Ciência e Tecnologia (SFRH/ BD/ 27528/ 2006) and by FCT project CMU-PT/HuMach/0039/2008 and by FCT (INESC-ID multi-annual funding) through the PIDDAC Program funds. Kuzman Ganchev was partially supported by NSF ITR EIA 0205448. Ben Taskar was partially supported by the DARPA CSSG 2009 Award and the ONR 2010 Young Investigator Award. Luísa Coheur was partially supported by FCT (INESC-ID multiannual funding) through the PIDDAC Program funds.

## Appendix A. Unsupervised Optimization

Berg-Kirkpatrick et al. (2010) describe the feature-rich HMM and show that training this model using direct gradient rather than EM can lead to better results. However, they only report results for the En45 corpus. Table 5 compares their implementation of both training regimes (BK+EM, BK+DG) on the different languages. Comparing the two training regimes, we see that there is no clear winner. BK+EM wins in 3 cases (Bg,En17,Dk) and loses on the other three.

It is also not clear how to predict which method is more suitable. In a follow up discussion [6] the authors propose that the difference arises from when each algorithm starts to fine-tune the weights of rare features relative to when it trains the weights of common features such as short suffixes. In the case of direct gradient training, at the start of optimization, the weights of common features change more rapidly because weight gradient is proportional to feature frequency. As training progresses, more weight is transferred to the rarer features. In contrast, for EM training, the optimization is done to completion on each M-Step, so even in the first iterations of EM where the counts are mostly random, the rarer features get a lot of the weight mass. This prevents the model from generalizing, and optimization terminates at a local maximum closer to the starting point. To allow EM to use common features for longer we tried some small experiments where we initially had very permissive stopping criteria for the M-step. After a few EM iterations with permissive stopping criteria, we require stricter stopping criteria. This tended to improve EM, but we did not find a principled method of setting a schedule for the convergence criteria on the M-step. Furthermore, these small experiments do not explain why direct gradient is only better than EM for some languages while being worse on others.

A related study (Salakhutdinov et al., 2003) compares the convergence rate of EM and direct gradient training, and identifies conditions when EM achieves Newton-like behavior, and when it achieves first-order convergence. The conditions are based on the amount of missing information, which in this case can be approximated by the number of hidden states. Potentially, this difference can also lead to different local maxima, mainly due to the non-local nature of the line search procedure of gradient based methods. In fact, looking at the results, DG training seems to work better on the corpora that have a higher number of hidden states (En45, Es) and work worse on corpora with fewer hidden states (Bg,En17).

Also in Table 5 we compare our implementation of the HMM+ME model to the implementation of Berg-Kirkpatrick et al. (2010), using the same conditions (regularization parameter, feature set, convergence criteria, initialization) and observe significant differences in results. Communication and code-comparison revealed small implementation differences: we use a bias feature while they do not; for the same random seed, our parameters are initialized differently than theirs; we have different implementations of the optimization algorithm; and a different number of iterations. For some corpora these differences result in better performance for their implementation, while for other corpora our implementation gets better results. We leave these details as well as a better understanding of the differences between each optimization procedure as future work, since this is not the main focus of the present paper.

---

6. `http://www.cs.berkeley.edu/~tberg/gradVsEM/main.html`

| | 1-Many | | | | | | 1-1 | | | | | |
|---|---|---|---|---|---|---|---|---|---|---|---|---|
| | En45 | En17 | Pt | Bg | Dk | Es | En45 | En17 | Pt | Bg | Dk | Es |
| BK+EM | 69.1 | 72.1 | 72.3 | **64.3** | 62.8 | 72.0 | 48.3 | **54.4** | 45.5 | **50.6** | 41.5 | 37.2 |
| BK+DG | **75.8** | 67.9 | **72.5** | 56.1 | 60.6 | **73.7** | **54.5** | 47.9 | 42.9 | 38.8 | 41.5 | 40.4 |
| HMM+ME | 67.1 | **72.2** | 72.3 | 61.1 | **65.1** | 71.8 | 45.0 | 51.1 | **46.3** | 46.1 | **42.6** | **40.8** |

| | VI | | | | | | V | | | | | |
|---|---|---|---|---|---|---|---|---|---|---|---|---|
| | En45 | En17 | Pt | Bg | Dk | Es | En45 | En17 | Pt | Bg | Dk | Es |
| BK+EM | 3.31 | **3.04** | **3.19** | **3.30** | 3.90 | 4.15 | .645 | **.568** | .582 | **.479** | .477 | .596 |
| BK+DG | **3.01** | 3.21 | 3.29 | 3.89 | 4.15 | **3.56** | **.678** | .534 | .574 | .392 | .477 | **.611** |
| HMM+ME | 3.77 | 3.11 | 3.21 | 3.46 | **3.77** | **3.56** | .606 | .564 | **.583** | .460 | **.519** | .608 |

Table 5: EM *vs* direct gradient from Berg-Kirkpatrick et al. (2010) implementation compared with our implementaion of EM of the HMM with maximum-entropy emission probabilities. The rows starting with BK are for the Berkeley implementation, while the rows starting with ME are for our implementation.

## Appendix B. Evaluation Metrics

To compare the performance of the different models one needs to evaluate the quality of the induced clusters. Several evaluation metrics for clustering have been proposed in previous work. The metrics we use to evaluate can be divided into two types (Reichart & Rappoport, 2009): mapping-based and information theoretic. Mapping based metrics require a post-processing step to map each cluster to a POS tag and then evaluate accuracy as for supervised POS tagging. Information-theoretic (IT) metrics compare the induced clusters directly with the true POS tags.

**1-Many** mapping and **1-1** mapping (Haghighi & Klein, 2006) are two widely-used mapping metrics. In the **1-Many** mapping, each hidden state is mapped to the tag with which it cooccurs the most. This means that several hidden states can be mapped to the same tag, and some tags might not be used at all. The **1-1** mapping greedily assigns each hidden state to a single tag. In the case where the number of tags and hidden states is the same, this will give a 1-1 correspondence. A major drawback of the latter mapping is that it fails to express all the information of the hidden states. Typically, unsupervised models prefer to explain very frequent tags with several hidden states, and combine some very rare tags. For example the Pt corpus has 3 tags that occur only once in the corpus. Grouping these together but subdividing nouns still provides a lot of information about the true tag assignments. However, this would not be captured by the **1-1** mapping. This metric tends to favor systems that produce an exponential distribution on the size of each induced cluster independent of the clusters' true quality, and it does not correlate well with the information theoretic metrics (Graça et al., 2009). Nevertheless, the 1-Many mapping also has drawbacks, since it can only distinguish clusters based on their most frequent tag. So, having a cluster split almost evenly between nouns and adjectives, or having a cluster with the same number of nouns, but a mixture of words with different tags gives the same 1-Many accuracy.

The information-theoretic measures we use for evaluation are variation of information (VI) (Meilă, 2007) and validity-measure (V) (Rosenberg & Hirschberg, 2007). Both are based on the entropy and conditional entropy of the tags and induced clusters. VI has desirable geometric properties – it is a metric and is convexly additive (Meilă, 2007). However, the range of VI values is dataset-dependent (VI lies in $[0, 2 \log N]$ where $N$ is the number of POS tags) which does not allow a comparison across datasets with different $N$. The validity-measure (V) is also an entropy-based measure and always lies in the range $[0, 1]$, but does not satisfy the same geometric properties as VI. It has been reported to give a high score when a large number of clusters exist, even if these are of low quality (Reichart & Rappoport, 2009). Other information-theoretic measures have been proposed that better handle different numbers of clusters, for instance NVI (Reichart & Rappoport, 2009). However, in this work all testing conditions will be on the same corpora with the same number of clusters so that problem does not exist. Christodoulopoulos, Goldwater, and Steedman (2010) present an extensive comparison between evaluation metrics. In related work Maron, Lamar, and Bienenstock (2010) present another empirical study about metrics and conclude that the VI metric can produce results that contradict the true quality of the induced clustering, by giving very high scores to very simple baseline systems, for instance assigning the same label to all words. They also point out several problems with the 1-1 metric some of which we explained previously. Since metric comparison is not the focus of this work we will compare all methods using the four metrics described in this section.

## References

Abeillé, A. (2003). *Treebanks: Building and Using Parsed Corpora*. Springer.

Afonso, S., Bick, E., Haber, R., & Santos, D. (2002). Floresta Sinta(c)tica: a treebank for Portuguese. In *Proc. LREC*, pp. 1698–1703.

Baum, L., Petrie, T., Soules, G., & Weiss, N. (1970). A maximization technique occurring in the statistical analysis of probabilistic functions of Markov chains. *The Annals of Mathematical Statistics*, *41*(1), 164–171.

Berg-Kirkpatrick, T., Bouchard-Côté, A., DeNero, J., & Klein, D. (2010). Painless unsupervised learning with features. In *Proc. NAACL*.

Bertsekas, D., Homer, M., Logan, D., & Patek, S. (1995). *Nonlinear programming*. Athena Scientific.

Brown, P. F., deSouza, P. V., Mercer, R. L., Pietra, V. J. D., & Lai, J. C. (1992). Class-based n-gram models of natural language. *Computational Linguistics*, *18*, 467–479.

Chen, S. (2003). Conditional and joint models for grapheme-to-phoneme conversion. In *Proc. ECSCT*.

Christodoulopoulos, C., Goldwater, S., & Steedman, M. (2010). Two decades of unsupervised POS induction: How far have we come?. In *Proc. EMNLP*, Cambridge, MA.

Civit, M., & Martí, M. (2004). Building cast3lb: A spanish treebank. *Research on Language & Computation*, *2*(4), 549–574.

Clark, A. (2003). Combining distributional and morphological information for part of speech induction. In *Proc. EACL*.

Dempster, A., Laird, N., & Rubin, D. (1977). Maximum likelihood from incomplete data via the EM algorithm. *Journal of the Royal Statistical Society. Series B (Methodological)*, *39*(1).

Freitag, D. (2004). Toward unsupervised whole-corpus tagging. In *Proc. COLING*. Association for Computational Linguistics.

Ganchev, K., Graça, J., Gillenwater, J., & Taskar, B. (2010). Posterior regularization for structured latent variable models. *Journal of Machine Learning Research*, *11*, 2001–2049.

Gao, J., & Johnson, M. (2008). A comparison of Bayesian estimators for unsupervised hidden Markov model POS taggers. In *In Proc. EMNLP*, pp. 344–352, Honolulu, Hawaii. ACL.

Goldwater, S., & Griffiths, T. (2007). A fully Bayesian approach to unsupervised part-of-speech tagging. In *In Proc. ACL*, Vol. 45, p. 744.

Graça, J., Ganchev, K., Pereira, F., & Taskar, B. (2009). Parameter vs. posterior sparisty in latent variable models. In *Proc. NIPS*.

Graça, J., Ganchev, K., & Taskar, B. (2007). Expectation maximization and posterior constraints. In *In Proc. NIPS*. MIT Press.

Graça, J. a. d. A. V. (2010). *Posterior Regularization Framework: Learning Tractable Models with Intractable Constraints*. Ph.D. thesis, Universidade Técnica de Lisboa, Instituto Superior Técnico.

Haghighi, A., & Klein, D. (2006). Prototype-driven learning for sequence models. In *Proc. HTL-NAACL*. ACL.

Headden, III, W. P., McClosky, D., & Charniak, E. (2008). Evaluating unsupervised part-of-speech tagging for grammar induction. In *Proc. COLING*, pp. 329–336.

Hwa, R., Resnik, P., Weinberg, A., Cabezas, C., & Kolak, O. (2005). Bootstrapping parsers via syntactic projection across parallel texts. *Special Issue of the Journal of Natural Language Engineering on Parallel Texts*, *11*(3), 311–325.

Johnson, M. (2007). Why doesn't EM find good HMM POS-taggers. In *In Proc. EMNLP-CoNLL*.

Kromann, Matthias T. (2003). The Danish Dependency Treebank and the underlying linguistic theory. In *Second Workshop on Treebanks and Linguistic Theories (TLT)*, pp. 217–220, Växjö, Sweden.

Lamar, M., Maron, Y., & Bienenstock, E. (2010). Latent-descriptor clustering for unsupervised POS induction. In *Proceedings of the 2010 Conference on Empirical Methods in Natural Language Processing*, pp. 799–809, Cambridge, MA. Association for Computational Linguistics.

Lamar, M., Maron, Y., Johnson, M., & Bienenstock, E. (2010). SVD and clustering for unsupervised POS tagging. In *Proceedings of the ACL 2010 Conference: Short Papers*, pp. 215–219, Uppsala, Sweden. Association for Computational Linguistics.

Lee, Y. K., Haghighi, A., & Barzilay, R. (2010). Simple type-level unsupervised POS tagging. In *Proceedings of the 2010 Conference on Empirical Methods in Natural Language Processing*, pp. 853–861, Cambridge, MA. Association for Computational Linguistics.

Marcus, M., Marcinkiewicz, M., & Santorini, B. (1993). Building a large annotated corpus of English: The Penn Treebank. *Computational linguistics*, *19*(2), 313–330.

Maron, Y., Lamar, M., & Bienenstock, E. (2010). Evaluation criteria for unsupervised POS induction. Tech. rep., Indiana University.

Martin, S., Liermann, J., & Ney, H. (1998). Algorithms for bigram and trigram word clustering. In *Speech Communication*, pp. 1253–1256.

Meilă, M. (2007). Comparing clusterings—an information based distance. *J. Multivar. Anal.*, *98*(5), 873–895.

Merialdo, B. (1994). Tagging English text with a probabilistic model. *Computational linguistics*, *20*(2), 155–171.

Moon, T., Erk, K., & Baldridge, J. (2010). Crouching Dirichlet, hidden Markov model: Unsupervised POS tagging with context local tag generation. In *Proc. EMNLP*, Cambridge, MA.

Neal, R. M., & Hinton, G. E. (1998). A new view of the EM algorithm that justifies incremental, sparse and other variants. In Jordan, M. I. (Ed.), *Learning in Graphical Models*, pp. 355–368. Kluwer.

Nocedal, J., & Wright, S. J. (1999). *Numerical optimization*. Springer.

Ratnaparkhi, A. (1996). A maximum entropy model for part-of-speech tagging. In *Proc. EMNLP*. ACL.

Ravi, S., & Knight, K. (2009). Minimized models for unsupervised part-of-speech tagging. In *In Proc. ACL*.

Reichart, R., & Rappoport, A. (2009). The NVI clustering evaluation measure. In *Proc. CONLL*.

Rosenberg, A., & Hirschberg, J. (2007). V-measure: A conditional entropy-based external cluster evaluation measure. In *EMNLP-CoNLL*, pp. 410–420.

Salakhutdinov, R., Roweis, S., & Ghahramani, Z. (2003). Optimization with EM and expectation-conjugate-gradient. In *Proc. ICML*, Vol. 20.

Schütze, H. (1995). Distributional part-of-speech tagging. In *Proc. EACL*, pp. 141–148.

Shen, L., Satta, G., & Joshi, A. (2007). Guided learning for bidirectional sequence classification. In *Proc. ACL*, Prague, Czech Republic.

Simov, K., Osenova, P., Slavcheva, M., Kolkovska, S., Balabanova, E., Doikoff, D., Ivanova, K., Simov, A., Simov, E., & Kouylekov, M. (2002). Building a Linguistically Interpreted Corpus of Bulgarian: the BulTreeBank. In *Proc. LREC*.

Smith, N., & Eisner, J. (2005). Contrastive estimation: Training log-linear models on unlabeled data. In *Proc. ACL*. ACL.

Snyder, B., Naseem, T., Eisenstein, J., & Barzilay, R. (2008). Unsupervised multilingual learning for POS tagging. In *Proceedings of the Conference on Empirical Methods in Natural Language Processing*, pp. 1041–1050. Association for Computational Linguistics.

Toutanova, K., & Johnson, M. (2007). A Bayesian LDA-based model for semi-supervised part-of-speech tagging. *In Proc. NIPS*, *20*.

Toutanova, K., Klein, D., Manning, C., & Singer, Y. (2003). Feature-rich part-of-speech tagging with a cyclic dependency network. In *In Proc. HLT-NAACL*.

Zhao, Q., & Marcus, M. (2009). A simple unsupervised learner for POS disambiguation rules given only a minimal lexicon. In *Proc. EMNLP*.